\PassOptionsToPackage{dvipsnames}{xcolor}
\documentclass{article}

\usepackage{microtype}
\usepackage{graphicx}
\usepackage{subcaption}
\usepackage{booktabs} 

\usepackage[dvipsnames, table]{xcolor}
\usepackage{hyperref}
\usepackage{fontawesome5}
\usepackage{pgfplots}
\pgfplotsset{compat=1.18}

\usepackage[preprint]{icml2026}

\usepackage{amsmath}
\usepackage{amssymb}
\usepackage{mathtools}
\usepackage{amsthm}
\usepackage{array}
\usepackage{tikz}
\usetikzlibrary{arrows.meta, positioning, fit, backgrounds, calc}
\usepackage[most]{tcolorbox}
\tcbuselibrary{breakable}
\usepackage{listings}
\usepackage{marginnote}
\usepackage{ragged2e}
\reversemarginpar 

\usepackage[capitalize,noabbrev]{cleveref}

\theoremstyle{plain}

\theoremstyle{definition}

\theoremstyle{remark}

\usepackage[most]{tcolorbox}
\usepackage{makecell}

\definecolor{judgeblue}{RGB}{225,238,255}
\definecolor{judgeframe}{RGB}{0,75,150}
\definecolor{judgelight}{RGB}{245,248,252}
\definecolor{judgebg}{RGB}{247,250,255}
\definecolor{judgeborder}{RGB}{0,70,140}

\usepackage[textsize=tiny]{todonotes}
\usepackage{amssymb}
\usepackage{pifont}
\usepackage{booktabs}   
\usepackage{array}      
\usepackage{ragged2e}   
\usepackage{tabularx}
\usepackage{colortbl}

\definecolor{darkblue}{RGB}{0,0,120}
\newcommand{\metric}[1]{\textcolor{darkblue}{\textbf{#1}}}
\definecolor{headerblue}{RGB}{230,242,255}
\definecolor{rowgray}{RGB}{248,248,248}
\definecolor{highlightrow}{RGB}{220,235,255}
\definecolor{metricblue}{RGB}{0,70,140}

\newcommand{\badval}[1]{\begingroup\setlength{\fboxsep}{1.2pt}\colorbox{red!18}{\textcolor{red!70!black}{#1}}\endgroup}
\newcommand{\goodval}[1]{\begingroup\setlength{\fboxsep}{1.2pt}\colorbox{green!15}{\textcolor{green!55!black}{#1}}\endgroup}

\newcommand{\codecell}[1]{%
{\ttfamily\footnotesize
\begin{tabular}[t]{@{}l@{}}#1\end{tabular}}}

\definecolor{softblue}{RGB}{170,200,255}

\newcommand{\colheat}[2]{\cellcolor{softblue!#1}#2}

\definecolor{softgreen}{RGB}{170,230,170}

\newcommand{\greenheat}[2]{\cellcolor{softgreen!#1}#2}

\newcolumntype{C}{>{\centering\arraybackslash}p{1.1cm}}
\tcbset{
    enhanced,
    colback=gray!3,
    colframe=black!80,
    boxrule=0.6pt,
    arc=2pt,
    left=6pt,
    right=6pt,
    top=6pt,
    bottom=6pt
}

\icmltitlerunning{When the API Speaks the Wrong Language: Revisiting Post-Training for Multilingual Tool Use}

\begin{document}

\twocolumn[
  \icmltitle{When the API Speaks the Wrong Language: Revisiting Post-Training for Multilingual Tool Use}



  \icmlsetsymbol{equal}{*}

  \begin{icmlauthorlist}
    \icmlauthor{Siddharth Chauhan}{comp}
    \icmlauthor{Thomas Butler}{comp}
    \icmlauthor{Abhishek Singhania}{comp}
    \icmlauthor{Pankaj Porwal}{comp}
    \icmlauthor{Honey Gupta}{comp}
  \end{icmlauthorlist}

  \icmlaffiliation{comp}{Amazon}

  \icmlcorrespondingauthor{Honey Gupta}{ghoney@amazon.com}

  \icmlkeywords{Machine Learning, ICML}

  \vskip 0.3in
]



\printAffiliationsAndNotice{}  

\begin{abstract}
The reliability of Large Language Models (LLMs) for API calling degrades in multilingual settings. A common failure occurs when a model selects the correct tool but generates argument values in an inconsistent language, which we term \textit{Argument Language Mismatch} (ALM). Although semantically correct, such outputs are operationally invalid and not captured by standard API-calling metrics.
We revisit post-training strategies for mitigating ALM and find that, in our benchmark, supervised fine-tuning (SFT) provides a strong baseline, substantially improving argument language consistency and end-to-end function call accuracy. Under consistent model selection, SFT achieves performance comparable to, and sometimes exceeding more complex reinforcement learning (RL) approaches.
We further examine whether RL with structured, argument-aware rewards offers additional benefits. While methods such as Group Relative Policy Optimization (GRPO) can improve language consistency and better preserve general reasoning ability, these gains are incremental and most pronounced in generalization and multi-objective trade-offs.
Overall, our results suggest that much of the performance in multilingual API grounding can be achieved through careful supervised training, with RL providing targeted rather than fundamental improvements.

\end{abstract}

\section{Introduction}

Large language models (LLMs) increasingly interact with external systems via structured API calls \citep{brown2020language,devlin2019bert}. In such settings, correctness requires not only selecting the appropriate API but also generating precisely formatted argument values. While recent advances have significantly improved tool-use capabilities in English, reliability degrades sharply in multilingual scenarios \citep{hu2023multi3woz}.

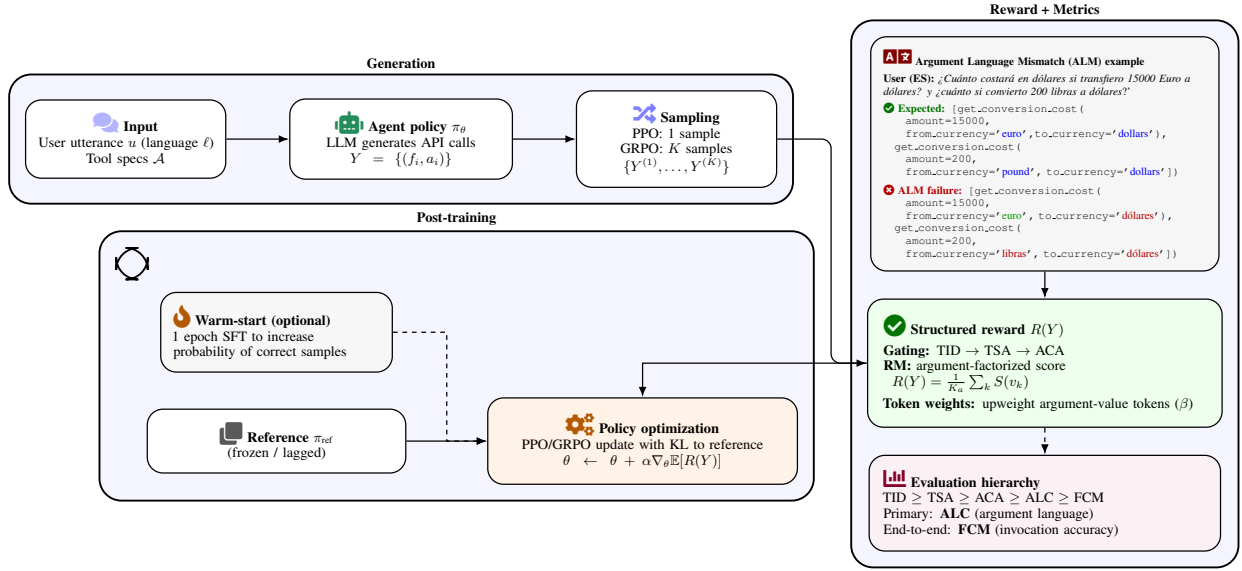
\begin{figure*}[t]
\centering
\resizebox{.95\textwidth}{!}{%
\begin{tikzpicture}[
  font=\small,
  stage/.style={draw, rounded corners=14pt, very thick, inner sep=10pt, fill=blue!4},
  box/.style={draw, rounded corners=10pt, thick, fill=white, inner sep=8pt, align=center},
  wide/.style={box, text width=5.2cm},
  med/.style={box, text width=4.4cm},
  sm/.style={box, text width=3.9cm},
  reward/.style={draw, rounded corners=10pt, thick, fill=green!7, inner sep=10pt, align=left, text width=7.2cm},
  metric/.style={draw, rounded corners=10pt, thick, fill=purple!6, inner sep=9pt, align=left, text width=7.2cm},
  inset/.style={draw, rounded corners=10pt, thick, fill=gray!7, inner sep=8pt, align=left, font=\scriptsize, text width=7.2cm},
  opt/.style={draw, rounded corners=10pt, thick, fill=gray!6, inner sep=8pt, align=left, text width=4.6cm},
  emph/.style={draw, rounded corners=10pt, thick, fill=orange!10, inner sep=10pt, align=center, text width=6.2cm},
  arrow/.style={-Latex, thick},
  darrow/.style={-Latex, thick, dashed},
  node distance=10mm and 14mm
]

\node[sm] (input) {
{\color{blue!30!white}\Large\faComments} \textbf{Input}\\
User utterance $u$ (language $\ell$)\\
Tool specs $\mathcal{A}$
};

\node[med, right=of input] (policy) {
{\color{Green!70!black}\Large\faRobot}\ \textbf{Agent policy} $\pi_\theta$\\
LLM generates API calls\\
$Y=\{(f_i, a_i)\}$
};

\node[sm, right=of policy] (sampling) {
{\color{blue!50!white}\Large\faRandom}\ \textbf{Sampling}\\
PPO: 1 sample\\
GRPO: $K$ samples\\
$\{Y^{(1)},\dots,Y^{(K)}\}$
};

\draw[arrow] (input) -- (policy);
\draw[arrow] (policy) -- (sampling);

\node[reward, right=20mm of sampling, yshift=-50mm] (rm) {
{\color{green!45!black}\Large\faCheckCircle}\ \textbf{Structured reward $R(Y)$}\\[1mm]
\textbf{Gating:} TID $\rightarrow$ TSA $\rightarrow$ ACA\\
\textbf{RM:} argument-factorized score\\
\hspace{2mm}$R(Y)=\frac{1}{K_a}\sum_k S(v_k)$\\[1mm]
\textbf{Token weights:} upweight argument-value tokens ($\beta$)
};

\node[inset, above=6mm of rm] (alm) {
{\color{red!50!black}\Large\faLanguage} \textbf{Argument Language Mismatch (ALM) example}\\[1mm]
\textbf{User (ES):} \textit{¿Cuánto costará en dólares si transfiero 15000 Euro a dólares? y ¿cuánto si convierto 200 libras a dólares}?'\\[1mm]
{\color{green!45!black}\faCheckCircle}\ \textcolor{green!45!black}{\textbf{Expected:}}
\texttt{[get\_conversion\_cost(}\\
\quad\quad\texttt{amount=15000,}\\
\quad\quad\texttt{from\_currency='}\textcolor{blue}{euro}\texttt{',to\_currency='}\textcolor{blue}{dollars}\texttt{'),}\\
\quad\texttt{get\_conversion\_cost(}\\
\quad\quad\texttt{amount=200,}\\
\quad\quad\texttt{from\_currency='}\textcolor{blue}{pound}\texttt{', to\_currency='}\textcolor{blue}{dollars}\texttt{'])}\\[4pt]
{\color{red!70!black}\faTimesCircle}\ \textcolor{red!70!black}
{\textbf{ALM failure:}}
\texttt{[get\_conversion\_cost(}\\
\quad\quad\texttt{amount=15000,}\\
\quad\quad\texttt{from\_currency='}\textcolor{green!60!black}{euro}\texttt{', to\_currency='}\textcolor{red!70!black}{d\'olares}\texttt{'),}\\
\quad\texttt{get\_conversion\_cost(}\\
\quad\quad\texttt{amount=200,}\\
\quad\quad\texttt{from\_currency='}\textcolor{red!70!black}{libras}\texttt{', to\_currency='}\textcolor{red!70!black}{d\'olares}\texttt{'])}\\
};

\node[metric, below=6mm of rm] (metrics) {
{\color{purple!70!black}\Large\faChartBar}\ \textbf{Evaluation hierarchy}\\
TID $\geq$ TSA $\geq$ ACA $\geq$ ALC $\geq$ FCM\\
Primary: \textbf{ALC} (argument language)  \\
End-to-end: \textbf{FCM} (invocation accuracy)
};

\draw[-Latex, thick, rounded corners=6pt]
(sampling.east) -- ++(12mm,0)
           -- ++(0, -50mm)
           -- (rm.west);

\draw[arrow] (alm.south) -- (rm.north);
\draw[darrow] (rm.south) -- (metrics.north);

\node[stage, below=15mm of policy, anchor=north, xshift=-60mm, ] (posttrain) {};

\node[opt, anchor=north west] (warm) at ([xshift=10mm,yshift=-10mm]posttrain.north west) {
{\color{orange!70!black}\Large\faFire}\ \textbf{Warm-start (optional)}\\
1 epoch SFT to increase\\
probability of correct samples
};

\node[wide, below=8mm of warm, anchor=north] (ref) {
{\color{gray!70!black}\Large\faClone}\ \textbf{Reference} $\pi_{\text{ref}}$\\
(frozen / lagged)
};

\node[emph, right=18mm of ref] (update) {
{\color{orange!70!black}\Large\faCogs}\ \textbf{Policy optimization}\\
PPO/GRPO update with KL to reference\\
$\theta \leftarrow \theta + \alpha \nabla_\theta \mathbb{E}[R(Y)]$
};

\draw[arrow] (ref) -- (update);

\draw[arrow] (rm.west) -- ++(0, 0mm) -| (update.north);

\draw[darrow] (warm.east) -- ++(12mm,0) |- (update.west);

\begin{scope}[on background layer]
  \node[stage, fit=(input) (policy) (sampling),
        label={[font=\small\bfseries]above:Generation}] {};
  \node[stage, fit=(alm) (rm) (metrics),
        label={[font=\small\bfseries]above:Reward + Metrics}] {};
  \node[stage, fit=(posttrain) (warm) (ref) (update),
        label={[font=\small\bfseries]above:Post-training}] {};
\end{scope}

\end{tikzpicture}%
}

\caption{
\textbf{Overview of our language-consistent API grounding framework.} Given a multilingual user request and available tool specifications, the model generates candidate API calls. A structured reward evaluates both structural correctness and whether argument values match the language of the user input. Reinforcement learning with argument-aware rewards encourages the model to produce language-consistent arguments, improving language consistency (ALC) and end-to-end function call accuracy~(FCM).
\textit{Inset:} Example of Argument Language Mismatch~(ALM), where the model selects the correct tool but generates argument values in the wrong language.}
\label{fig:main_overview}
\end{figure*}

A common and underexplored failure mode arises when a model selects the correct API but generates argument values in a language inconsistent with the user input or system requirements. We refer to this phenomenon as \textit{Argument Language Mismatch} (ALM). Although such outputs are often semantically correct, they are operationally invalid in real-world systems that enforce strict language constraints, leading to complete task failure. 
ALM is not a semantic or intent-recognition error. As illustrated in Figure~\ref{fig:main_overview}, models frequently succeed at intent understanding and API selection, yet fail to condition argument realization on the user’s language.
Because standard API-calling metrics treat these cases as generic invocation errors, they obscure a major source of failure in multilingual agentic systems.


Existing approaches to multilingual modeling, including supervised fine-tuning (SFT) \citep{conneau2020xlmr, wang2022supernatinst, lin2021bitod, hu2023multi3woz}, have demonstrated strong performance in instruction-following and tool-use tasks. In particular, SFT can effectively improve argument-level language consistency when training and test distributions are aligned. However, it remains unclear whether such improvements extend robustly to more challenging settings, such as unseen APIs, diverse argument structures, or cross-lingual transfer. At the same time, recent work suggests that reinforcement learning (RL) with structured objectives can improve alignment in complex generation tasks \citep{ouyang2022training, xu2024dpo, gehring2024rlef}, raising the question of whether such methods are necessary for addressing ALM.

In this work, we visit post-training strategies for mitigating ALM in multilingual API calling. Rather than assuming that increasingly complex objectives are required, we first examine how far strong supervised baselines can go. Surprisingly, we find that supervised fine-tuning alone resolves a large fraction of ALM errors, yielding substantial improvements in both argument language consistency and end-to-end function call accuracy. Under controlled and consistent model selection, SFT achieves performance comparable to, and in some cases exceeding, more complex RL-based approaches.

We then investigate whether reinforcement learning with structured, argument-aware rewards provides additional benefits beyond SFT. To this end, we formulate multilingual API calling as a structured generation problem and design reward functions that explicitly evaluate argument language consistency. Our approach includes hierarchical rewards aligned with the API-calling process, argument-factorized credit assignment, and token-level reward weighting. We study both \textit{Proximal Policy Optimization} (PPO) \citep{schulman2017ppo} and \textit{Group Relative Policy Optimization} (GRPO) \citep{shao2024deepseekmath}, enabling a controlled comparison of optimization strategies under identical reward formulations. This setup allows us to isolate when and how RL contributes beyond supervised learning.

To support this study, we construct a multilingual extension of the Berkeley Function Calling benchmark~\citep{patil2024bfcl}, covering multiple languages while preserving realistic API structure and argument diversity. We evaluate models under both \textit{learnability} and \textit{generalization} settings, as well as cross-lingual transfer to unseen languages. Across these settings, we find that reinforcement learning provides incremental improvements over strong SFT baselines, with the most consistent gains observed in generalization and in preserving general reasoning ability.

In summary, this paper makes three key contributions:
\begin{itemize}
    \item We formalize \textit{Argument Language Mismatch} (ALM) as a distinct and practically important failure mode in multilingual API calling.
    \item We show that supervised fine-tuning (SFT) provides a strong baseline for mitigating ALM, achieving substantial improvements in language consistency and end-to-end accuracy.
    \item We evaluate reinforcement learning with structured, argument-aware rewards and find that it yields incremental gains over SFT, particularly in generalization and reasoning preservation.
    \item We provide a systematic comparison of post-training strategies, highlighting when additional training complexity is beneficial for structured multilingual generation.
\end{itemize}


\section{Problem Formulation}
\label{sec:prob}

We study multilingual API calling as a structured prediction problem in which a language model must map a natural language user utterance into a sequence of API invocations with correctly formatted arguments. Our focus is on isolating  \emph{Argument Language Mismatch (ALM)}, a failure mode that is not captured by standard API-calling metrics.

\subsection{Multilingual API Calling}

Each data point consists of a user request $u$ written in a natural language $\ell \in \mathcal{L}$ and a set of available API specifications $\mathcal{A}$. The model must generate a structured output $Y$ consisting $\geq 0$ API calls:
\[
Y = \{ (f_1, \mathbf{a}_1), (f_2, \mathbf{a}_2), \ldots, (f_m, \mathbf{a}_m) \},
\]
where each $f_i \in \mathcal{A}$ is an API name and $\mathbf{a}_i = \{ a_{i,1}, \ldots, a_{i,K_i} \}$ are its arguments. Each argument $a_{i,k}$ contains a value $v_{i,k}$ that is either categorical or free-form natural language.

In multilingual settings, argument values are not purely semantic objects: they also carry a \emph{language attribute}. Let $\mathrm{lang}(v)$ denote the language in which an argument value is expressed. In correctly grounded API calls, argument values that originate from the user must satisfy:
\[
\mathrm{lang}(v_{i,k}) = \mathrm{lang}(u),
\]
unless the API specification explicitly requires another language.

\subsection{Argument Language Mismatch}

\emph{Argument Language Mismatch (ALM)} occurs when the model selects the correct API and argument names, but produces argument values in the wrong language. Formally, for a generated output $Y$ and a ground-truth output $Y^{\star}$, an argument $a_{i,k}$ is said to exhibit ALM if
$\mathrm{lang}(v_{i,k}) \neq \mathrm{lang}(v^{\star}_{i,k})$,
where $v^{\star}_{i,k}$ is the ground-truth value. Figure~\ref{fig:main_overview} provides an example where a Spanish user request is correctly mapped to the food-ordering API, but its argument values are produced in English, causing a downstream system failure.

ALM is fundamentally different from semantic errors: a response can be semantically correct yet operationally invalid due to language inconsistency. Standard metrics such as AST matching or exact string match collapse these failures into a single error class, obscuring the underlying cause.

\subsection{Hierarchical Evaluation Metrics}
\label{sec:metrics}
To isolate \textit{Argument Language Mismatch} (ALM) from other sources of error in multilingual API calling, we evaluate model outputs using a hierarchical set of turn-level metrics. Each metric corresponds to a distinct stage of the tool invocation process and is evaluated \textit{conditionally on the success of all preceding stages}.
This decomposition follows evaluation practices in tool-use and semantic parsing benchmarks, where correctness is assessed separately for tool selection, argument realization, and full execution \citep{qin2023toollm, li2023apibank, guo2024stabletoolbench, zang2020multiwoz, yu2018spider}.

\paragraph{Tool Invocation Detection (TID).}
Whether the model correctly determines that an API call should be invoked for the current user turn.

\paragraph{Tool Selection Accuracy (TSA).}
Given that an API call is required, whether the model selects the correct API function(s).

\paragraph{Argument Completion Accuracy (ACA).}
Given correct tool selection, whether the model generates all required argument names specified by the API schema.

\paragraph{Argument Language Consistency (ALC).}
Given correct argument completion, whether all required textual argument values are expressed in the same language as the user input.

\paragraph{Function Call Match (FCM)}.
Whether the complete API call, including the function name and all argument values, matches the ground-truth invocation. Since exact-match metrics such as AST~\cite{patil2024bfcl} can be overly restrictive for free-form arguments, we instead compute accuracy using a combination of exact matching and semantic similarity.

These metrics form a strict hierarchy:
\[
\text{FCM} \leq \text{ALC} \leq \text{ACA} \leq \text{TSA} \leq \text{TID}.
\]
allowing us to distinguish between failures due to API selection, argument omission, language mismatch, and full semantic errors. ALC is analogous to slot-value accuracy in task-oriented dialogue, but specialized to measure language consistency of argument values. Our primary objective is to maximize ALC without sacrificing FCM or general reasoning ability.

\subsection{Learning Objective}

Let $\pi_\theta(Y \mid X)$ denote the model policy, where $X = (u, \mathcal{A})$ is the input. Our goal is to learn a policy that maximizes expected task reward:
\[
\max_\theta \; \mathbb{E}_{Y \sim \pi_\theta(\cdot \mid X)} [ R(Y) ],
\]
where $R(Y)$ is a structured reward function designed to provide explicit credit for producing language-consistent argument values. While ALM appears to require structured objectives, we investigate whether standard supervised learning already suffices to mitigate this failure mode and how it compares to reinforcement learning approaches that explicitly optimize for language consistency.









\section{Dataset and Benchmark Construction}
\label{sec:dataset}

Studying multilingual API grounding requires datasets that simultaneously exhibit rich tool-use structure and cross-lingual coverage. Existing benchmarks typically satisfy only one of these requirements: multilingual dialogue datasets provide strong language coverage but limited tool complexity, while function-calling datasets offer realistic tool schemas but are predominantly English.

\subsection{Dataset Selection}

To identify a benchmark suitable for studying Argument Language Mismatch (ALM), we evaluated several API-calling datasets across dimensions of tool-use complexity, including the number of function calls per turn, arguments per API, and the presence of non-categorical argument values.

Multilingual dialogue datasets such as MULTI3WOZ \citep{zang2020multiwoz} and BiToD \citep{lin2021bitod} provide strong cross-lingual coverage but rely on slot-filling schemas with limited argument diversity and minimal tool-selection complexity. Consequently, they rarely expose failures where models must generate free-form argument values conditioned on user language.
In contrast, recent function-calling datasets including APIGen \citep{liu2024apigen}, ToolAlpaca \citep{guo2024stabletoolbench}, ToolBench \citep{qin2023toolllm, qin2023toollm, guo2024stabletoolbench}, API-Bank \citep{li2023apibank}, and Glaive FC v2 \citep{glaive_function_calling_v2} provide richer API schemas but are largely English-only and often synthetically generated.

Based on this analysis, we select the Berkeley Function Calling (BFC) dataset \citep{patil2024bfcl}. BFC provides human-annotated supervision, multi-turn interactions, multiple candidate APIs per turn, and a high proportion of free-form argument values, making it well suited for studying argument-level language grounding errors such as ALM.

\subsection{Multilingual Benchmark Extension}

The original BFC dataset is predominantly English. To enable multilingual evaluation, we construct a translated version covering five languages: Spanish (Es), French (Fr), Italian (It), and Dutch (Nl).

In particular, some argument values represent user-provided natural language (e.g., food items, descriptions, or queries) and must be translated, while others correspond to canonical identifiers, named entities, or API-enforced formats that must remain unchanged. We therefore design a rule-based translation protocol guided by the API specifications, ensuring coherence between the translated user utterances and the resulting argument values.
The resulting corpus forms a parallel multilingual benchmark, where each dialogue is available in all five languages.

\subsection{Data Splits}

To evaluate both memorization and generalization, we construct two evaluation splits from the translated dataset. First, we identify all turns that are relevant to ALM by selecting those that satisfy at least one of the following conditions:
(i) at least one argument value is translated into a non-English language, or
(ii) the turn exhibits ALM under baseline model evaluation (e.g., Spanish expected but English predicted). This yields $832$ turns ($16.35\%$ of the full dataset), which we use for all training and evaluation. From these examples we derive two complementary splits:

\paragraph{Split-1 (Learnability)}: moderate API overlap between training and test sets, measuring the ability to learn language-consistent argument generation when similar APIs appear during training.

\paragraph{Split-2 (Generalization)}: minimal API overlap between training and test sets, evaluating whether models learn language-matching rules that transfer to unseen APIs and argument structures.

\noindent All metrics are computed at the turn level rather than the dialogue level. Models are trained on Spanish only and evaluated on Spanish as well as unseen languages (Italian, Dutch, and French), enabling controlled evaluation of cross-lingual transfer.

\section{Learning Algorithms and Reward Models}
\label{sec:methods}

Our goal is to study how different post-training strategies mitigate Argument Language Mismatch (ALM) in multilingual API calling. Rather than assuming that complex objectives are necessary, we begin with supervised fine-tuning (SFT) as a baseline and systematically evaluate whether reinforcement learning (RL) provides additional benefits.


\subsection{Training Paradigms}

We consider two classes of post-training methods: supervised fine-tuning and reinforcement learning with structured rewards. This setup allows us to isolate the contribution of explicit optimization for argument-level language consistency beyond standard likelihood-based training.


\subsubsection{Supervised Fine-Tuning (SFT)}
We first consider supervised fine-tuning, which serves as our first approach for improving multilingual API grounding. Given an input $X = (u, A)$ consisting of a user utterance $u$ and a set of available APIs $A$, supervised fine-tuning maximizes the likelihood of the ground-truth structured output $Y^\star$:
\[
\mathcal{L}_{\text{SFT}} = - \mathbb{E}_{(X,Y^\star)} \sum_{t \in Y^\star} \log \pi_\theta(y_t \mid X, y_{<t}).
\]

This objective encourages the model to imitate language-consistent API calls observed in the training data. In multilingual settings, SFT implicitly learns to align argument values with the user’s language through exposure to parallel examples.

\subsubsection{Reinforcement Learning}
To evaluate whether explicit optimization for language consistency provides additional gains, we extend the baseline with reinforcement learning. Instead of imitating reference outputs, the model samples candidate API calls $Y \sim \pi_\theta(\cdot \mid X)$ and receives a structured reward that evaluates both structural correctness and language consistency. We study two policy optimization algorithms: Proximal Policy Optimization (PPO) and Group Relative Policy Optimization (GRPO). 

\paragraph{Proximal Policy Optimization (PPO)}
We first apply Proximal Policy Optimization (PPO), a widely used policy-gradient method for aligning language models. PPO samples a structured output $Y$ and updates the policy using a clipped policy-gradient objective with KL 
regularization \citep{schulman2017ppo}: \\

\resizebox{\linewidth}{!}{$
\displaystyle
\begin{aligned}
\mathcal{L}_{\text{PPO}}(\theta)
&=
- \mathbb{E}_{(X,Y)}
\Bigg[
\sum_{t \in Y}
\min \Big(
r_t(\theta)\, \hat{A}_t,\;
\text{clip}\!\left(
r_t(\theta),\, 1-\epsilon,\, 1+\epsilon
\right)\hat{A}_t
\Big)
\Bigg] \\
&\quad
+ \beta\, \mathbb{E}_{(X,Y)}
\big[
\mathrm{KL}(\pi_\theta \,\|\, \pi_{\text{ref}})
\big].
\end{aligned}
$}

Here, $r_t(\theta) = \frac{\pi_\theta(y_t \mid X, y_{<t})}{\pi_{\theta_{\text{old}}}(y_t \mid X, y_{<t})}$ is token-level importance ratio, $\hat{A}_t$ denotes the estimated advantage.

\paragraph{Group Relative Policy Optimization (GRPO)}
While PPO updates the policy based on a single sampled response, GRPO samples $K$ structured outputs $\{Y^{(j)}\}_{j=1}^K$ from the current policy and 
compares them using relative rewards. For each response $j$, the advantage is computed as
\[
\hat{A}^{(j)} = \frac{R^{(j)} - \mu_R}{\sigma_R + \delta},
\]
where $\mu_R$ and $\sigma_R$ are the mean and standard deviation of rewards within the sampled group.

The policy is then updated using a clipped policy-gradient objective:\\

\resizebox{.9\linewidth}{!}{$
\displaystyle
\begin{aligned}
\mathcal{L}_{\text{GRPO}}
&=
- \mathbb{E}_{X}
\Bigg[
\frac{1}{K}
\sum_{j=1}^K
\sum_{t \in Y^{(j)}}
\min \Big(
r_t^{(j)} \hat{A}^{(j)},\;
\text{clip}\!\left(
r_t^{(j)},\, 1-\epsilon,\, 1+\epsilon
\right)\hat{A}^{(j)}
\Big)
\Bigg] \\
&\quad
+ \beta\,
\mathrm{KL}.
\end{aligned}
$}

\noindent Because GRPO compares multiple candidate responses for the same prompt, it encourages exploration over alternative argument realizations. 

\subsection{Reward Design for Language-Consistent API Grounding}
\label{sec:reward_models}

To mitigate Argument Language Mismatch (ALM), the reward function must explicitly evaluate whether generated argument values match the language of the user input. Standard supervised objectives provide token-level likelihood signals but do not directly capture argument-level language consistency. We therefore design reward functions that progressively increase the granularity of feedback provided to the policy.

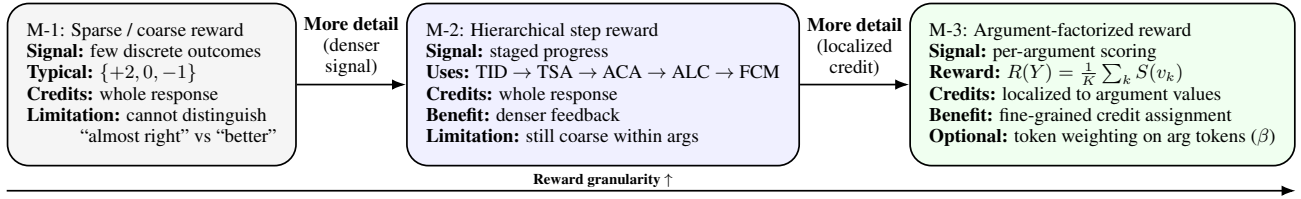
\begin{figure*}[t]
\centering
\resizebox{\textwidth}{!}{%
\begin{tikzpicture}[
    font=\small,
    box/.style={draw, rounded corners=10pt, thick, inner sep=9pt, align=left, fill=white},
    rm1/.style={box, fill=gray!8},
    rm2/.style={box, fill=blue!6},
    rm3/.style={box, fill=green!6},
    arrow/.style={-Latex, thick},
    label/.style={font=\bfseries},
]

\node[rm1] (rm1) {
{\label RM-1: Sparse / coarse reward}\\
\textbf{Signal:} few discrete outcomes\\
\textbf{Typical:} $\{+2, 0, -1\}$\\
\textbf{Credits:} whole response\\
\textbf{Limitation:} cannot distinguish\\
\hspace{8mm} ``almost right'' vs ``better''
};

\node[rm2, right=18mm of rm1] (rm2) {
{\label RM-2: Hierarchical step reward}\\
\textbf{Signal:} staged progress\\
\textbf{Uses:} TID $\rightarrow$ TSA $\rightarrow$ ACA $\rightarrow$ ALC $\rightarrow$ FCM\\
\textbf{Credits:} whole response\\
\textbf{Benefit:} denser feedback\\
\textbf{Limitation:} still coarse within args
};

\node[rm3, right=18mm of rm2] (rm3) {
{\label RM-3: Argument-factorized reward}\\
\textbf{Signal:} per-argument scoring\\
\textbf{Reward:} $R(Y)=\frac{1}{K}\sum_k S(v_k)$\\
\textbf{Credits:} localized to argument values\\
\textbf{Benefit:} fine-grained credit assignment\\
\textbf{Optional:} token weighting on arg tokens ($\beta$)
};

\draw[arrow] (rm1) -- node[above, align=center] {\textbf{More detail}\\(denser \\ signal)} (rm2);
\draw[arrow] (rm2) -- node[above, align=center] {\textbf{More detail}\\(localized \\ credit)} (rm3);

\draw[arrow] ([yshift=-4mm]rm1.south west) -- ([yshift=-4mm]rm3.south east);
\node[below=0mm of rm2, align=center, font=\scriptsize\bfseries] {Reward granularity $\uparrow$};

\end{tikzpicture}
}
\caption{
\textbf{Reward granularity increases from RM-1 to RM-3.}
RM-1 provides sparse, response-level feedback; RM-2 introduces hierarchy-aware step rewards;
RM-3 factorizes reward across argument values, enabling fine-grained credit assignment for mitigating Argument Language Mismatch (ALM).
}
\label{fig:rm_granularity}
\vspace{-.2cm}
\end{figure*}

Let $X=(u,A)$ denote the input consisting of a user utterance $u$ and available APIs $A$, and let $Y\sim\pi_\theta(\cdot\mid X)$ denote the generated API call sequence. Our goal is to design reward functions $R(Y)$ that reflect the hierarchical evaluation metrics introduced in Section 2.3 while providing informative training signals for reinforcement learning.



\subsubsection{RM-1: Sparse Binary Reward}
We first consider a minimal reward formulation that distinguishes only between fully correct outputs and different classes of failures.  RM-1 provides a coarse baseline reward that distinguishes only between perfect correctness, language-only failures, and structural errors:
\begin{equation}
R_{\text{RM-1}}(Y)=
\begin{cases}
+2.0 & \text{if } \text{FCM}(Y)=1,\\
0.0 & \text{if } \text{TID}=\text{TSA} \\
& =\text{ACA}=1
~\land \\
& \text{ALC}=0,\\
-1.0 & \text{otherwise}.
\end{cases}
\end{equation}
This reward collapses most non-perfect outputs into a small set of discrete outcomes. While simple, such sparse feedback provides limited guidance for correcting partial errors such as language mismatches.

\subsubsection{RM-2: Hierarchical Step Reward}
To provide denser feedback aligned with the API-grounding process, RM-2 assigns intermediate rewards according to the deepest level of the evaluation hierarchy achieved by a generated output.

\paragraph{Binary and Graded ALC}
For each required text argument $a_{i,k}$ with value $v_{i,k}$, an LLM judge assigns a language score
\[
s_{i,k} \in \{2.0,\,1.5,\,1.0\},
\]
corresponding to correct language, partial match, and language mismatch, respectively. The graded score is defined as the average over all required arguments,
\begin{equation}
\text{ALC}_{\text{cont}}(Y)=\frac{1}{K*I}\sum_{i,k}s_{i,k},
\end{equation}
which takes values in $[1.0, 2.0]$.

A binary language-consistency score is obtained by thresholding this value: $\text{ALC}=1$ if $\text{ALC}_{\text{cont}} \ge 1.8$ (i.e., at least $90\%$ of the maximum score 2.0), and $\text{ALC}=0$ otherwise. While the binary score determines whether language consistency is considered successful, the continuous score is used within RM-2 to distinguish between complete language failure ($\text{ALC}_{\text{cont}}\le1.0$) and partial success cases where at least one argument is expressed in the correct language ($\text{ALC}_{\text{cont}}>1.0$). \\

\resizebox{.85\linewidth}{!}{$
\displaystyle
R_{\text{RM-2}}(Y)=
\begin{cases}
-1.0 & \text{if } \text{TID}=0,\\
-0.5 & \text{if } \text{TID}=1,\,\text{TSA}=0,\\
+0.5 & \text{if } \text{TSA}=1,\,\text{ACA}=0,\\
+1.0 & \text{if } \text{ACA}=1,\,\text{ALC}_{\text{cont}}\le1.0,\\
+1.5 & \text{if } \text{ACA}=1,\,\text{ALC}_{\text{cont}}>1.0,\\
+2.0 & \text{if } \text{FCM}=1.
\end{cases}
$}

This reward structure separates structural correctness from language errors and provides more informative feedback than RM-1.

\subsubsection{RM-3: Argument-Factorized Reward}
While RM-2 improves learning by providing intermediate rewards, it still assigns identical rewards to outputs with different argument-level quality. To better align the training signal with the structure of the error, we introduce an argument-factorized reward.

For each argument value $v_{i,k}$, the judge assigns:
\begin{equation}
S(v_{i,k})\in\{2.5,\,2.0,\,1.0\},
\end{equation}
corresponding to exact language match, correct language with minor variation, and language mismatch. The graded argument-level score is defined as:
\begin{equation}
\text{ALC}_{\text{cont}}(Y)=\frac{1}{K}\sum_{i,k} S(v_{i,k}).
\end{equation}

RM-3 combines discrete structural gates with continuous argument-level rewards:

\resizebox{.9\linewidth}{!}{$
\displaystyle
R_{\text{RM-3}}(Y)=
\begin{cases}
-1.0 & \text{if } \text{TID}=0,\\
-0.5 & \text{if } \text{TID}=1,\,\text{TSA}=0,\\
+0.5 & \text{if } \text{TID}=1,\,\text{TSA}=1,\,\text{ACA}=0,\\
\text{ALC}_{\text{cont}}(Y) & \text{otherwise}.
\end{cases}
$}

\vspace{.2cm}
Unlike RM-1 and RM-2, RM-3 provides continuous feedback once structural correctness is satisfied, enabling fine-grained credit assignment aligned with individual argument realizations.

\begin{table}[t]
\centering
\caption{Two API calls that differ only in argument language. RM-3 assigns higher reward to the language-consistent version.}
\label{tab:reward_example}
\resizebox{\linewidth}{!}{
\begin{tabular}{lccc}
\toprule
Response & RM-1 & RM-2 & RM-3 \\
\midrule
\makecell{[g\textit{et\_conversion\_cost(}\\
amount=15000, from\_currency=\textcolor{green}{euro}, to\_currency=\textcolor{red}{dólares}),\\
\textit{get\_conversion\_cost(}\\
amount=200, from\_currency=\textcolor{red}{libras}, to\_currency=\textcolor{green}{dólares})]
} & 0 & 1.5 & 1.375 \\
\midrule
\makecell{[\textit{get\_conversion\_cost(}\\
amount=15000, from\_currency=\textcolor{green}{euro}, to\_currency=\textcolor{green}{dollars}), \\
\textit{get\_conversion\_cost(}\\
amount=200, from\_currency=\textcolor{red}{libras}, to\_currency=\textcolor{green}{dollars})]} & 0 & 1.5 & 2.125 \\
\bottomrule
\end{tabular}}
\end{table}


\subsection{Token-Level Reward Weighting}
In reinforcement learning, scalar rewards are typically distributed uniformly across all tokens. However, only argument-value tokens determine language consistency. To better align the training signal with the source of error, we introduce token-level reward weighting, assigning higher weights to tokens corresponding to argument values.

We study whether emphasizing these tokens improves learning efficiency and language consistency, particularly under fine-grained reward formulations.
We introduce a multiplicative weight $\beta$ for these tokens:
\[
R_{i,t} =
\begin{cases}
\beta \cdot R_i & \text{if token $t$ is in an argument value}\\
R_i & \text{otherwise}.
\end{cases}
\]
We study $\beta \in \{1.5,3\}$. This improves GRPO but destabilizes PPO due to batch-level normalization.

\subsection{SFT Warm-Start}

Because on-policy reinforcement learning depends on sampling high-quality candidates, we initialize RL training from a model obtained via supervised fine-tuning. We therefore perform one epoch of SFT before RL:
\[
\theta_0 \leftarrow \arg\min_\theta \mathcal{L}_{\text{SFT}}(\theta),
\quad
\theta \leftarrow \text{RL}(\theta_0).
\]
This warm-start increases the probability of generating structurally correct and partially language-consistent outputs, improving training stability and sample efficiency.



\section{Experiments and Results}
\label{sec:experiments}

We evaluate post-training strategies for mitigating Argument Language Mismatch (ALM) in multilingual API grounding. Rather than focusing solely on improving performance through increasingly complex methods, our goal is to understand how far strong supervised baselines can go, and when additional reinforcement learning (RL) objectives provide meaningful gains.


\subsection{Experimental Setup}

We conduct experiments on the multilingual extension of the Berkeley Function Calling (BFC) benchmark (Section~\ref{sec:dataset}). Models are trained on Spanish and evaluated on both Spanish and unseen languages (Italian, Dutch, and French), enabling controlled evaluation of cross-lingual transfer. 

We report results using the hierarchical metrics introduced in Section 2.3, including Tool Invocation Detection (TID), Tool Selection Accuracy (TSA), Argument Completion Accuracy (ACA), Argument Language Consistency (ALC), and Function Call Match (FCM). All metrics are computed at the turn level.

We compare supervised fine-tuning (SFT) with reinforcement learning methods (PPO and GRPO) under two evaluation protocols. In the \textbf{epoch-fixed setting}, all methods are trained for a comparable budget. In the \textbf{validation-selected setting}, checkpoints are selected based on validation FCM to ensure fair comparison across methods. {Epoch-fixed results are reported in Table~\ref{tab:epoch_fixed_main}, while the best-checkpoint (validation-selected) results, which are the focus of most of our analysis, are consolidated in Table~\ref{tab:selection_control}. Unless otherwise noted, the ablation studies (Tables~\ref{tab:reward_ablation}, \ref{tab:ppo_grpo}, and~\ref{tab:token_weight}) likewise report the best validation checkpoint.}





\paragraph{Base Models.}
We use the Qwen2.5 model family, including \emph{Qwen2.5-7B-Instruct},
\emph{Qwen2.5-14B-Instruct}, and \emph{Qwen2.5-32B-Instruct}.
Unless explicitly stated, results in this section use 14B model.

\subsection{Supervised Fine-Tuning for ALM}
We begin by evaluating SFT as a baseline for mitigating ALM. As shown in Table~\ref{tab:epoch_fixed_main}, across both learnability and generalization splits, SFT substantially improves argument language consistency (ALC) and end-to-end function call accuracy (FCM) over the base model.

Notably, SFT resolves a large fraction of ALM errors while maintaining high performance on tool invocation, selection, and argument completion. These results indicate that standard likelihood-based training already captures much of the structure required for language-consistent argument generation.

\subsection{Reinforcement Learning for ALM}

We next compare SFT with RL-based methods under the same training budget. As shown in Table~\ref{tab:epoch_fixed_main}, reinforcement learning, particularly GRPO, improves ALC and FCM relative to the base model, and also the gains over SFT are modest.

{More importantly, when both families are compared at their best validation checkpoint (Table~\ref{tab:selection_control}), the gap between SFT and RL further narrows, and in fact reverses on Split-1: best-checkpoint SFT reaches 79.1 ALC / 67.4 FCM, exceeding both GRPO (74.0 / 55.3) and SFT+GRPO (79.3 / 61.3) on FCM. In this setting, SFT achieves performance comparable to, and in some cases exceeding, RL-based methods. These results suggest that much of the improvement attributed to RL can be achieved through careful supervised training and model selection.}

\begin{table*}[t]
\centering
\small
\caption{Epoch-fixed comparison of post-training algorithms on Split-1 (learnability; high API overlap) and Split-2 (generalization; low API overlap).
All methods are initialized from the same base model (\texttt{Qwen2.5-14B-Instruct}).}
\setlength{\tabcolsep}{4.5pt}
\begin{tabular}{c@{\hspace{8pt}}c}
\begin{minipage}{0.92\textwidth}
\centering
\begin{tabular}{lccccc|ccccc}
\toprule
& \multicolumn{5}{c|}{\textbf{Split-1: Learnability}} 
& \multicolumn{5}{c}{\textbf{Split-2: Generalization}} \\
\cmidrule(lr){2-6}\cmidrule(lr){7-11}
\textbf{Method} 
& TID & TSA & ACA & ALC & FCM
& TID & TSA & ACA & ALC & FCM \\
\midrule
Base      
& \colheat{20}{99.57} & \colheat{10}{91.48} & \colheat{10}{91.48} & \colheat{0}{52.34} & \colheat{0}{32.34}
& \colheat{20}{99.72} & \colheat{20}{94.59} & \colheat{20}{94.59} & \colheat{0}{45.94} & \colheat{0}{22.16} \\

SFT       
& \colheat{20}{100.0} & \colheat{60}{94.04} & \colheat{60}{94.04} & \colheat{40}{63.82} & \colheat{25}{40.42}
& \colheat{20}{100.0} & \colheat{70}{95.40} & \colheat{70}{95.40} & \colheat{25}{54.34} & \colheat{15}{26.75} \\

GRPO      
& \colheat{20}{100.0} & \colheat{15}{91.91} & \colheat{15}{91.91} & \colheat{85}{74.47} & \colheat{80}{51.49}
& \colheat{20}{100.0} & \colheat{10}{93.51} & \colheat{10}{93.51} & \colheat{80}{69.73} & \colheat{75}{42.70} \\

SFT+GRPO  
& \colheat{20}{100.0} & \colheat{50}{93.61} & \colheat{50}{93.61} & \colheat{100}{75.32} & \colheat{100}{54.04}
& \colheat{20}{100.0} & \colheat{70}{95.40} & \colheat{70}{95.40} & \colheat{100}{72.70} & \colheat{100}{45.59} \\

PPO       
& \colheat{20}{100.0} & \colheat{30}{92.76} & \colheat{30}{92.76} & \colheat{70}{71.91} & \colheat{65}{48.36}
& \colheat{20}{99.46} & \colheat{100}{95.60} & \colheat{100}{95.60} & \colheat{40}{61.90} & \colheat{45}{37.84} \\

SFT+PPO   
& \colheat{20}{100.0} & \colheat{40}{93.19} & \colheat{40}{93.19} & \colheat{80}{73.61} & \colheat{75}{49.78}
& \colheat{20}{100.0} & \colheat{75}{95.41} & \colheat{75}{95.41} & \colheat{65}{68.11} & \colheat{65}{41.62} \\

\bottomrule
\end{tabular}
\end{minipage}
&
\begin{minipage}{0.04\textwidth}
\centering
\vspace{2pt}
\begin{tikzpicture}
  \shade[top color=softblue!120, bottom color=softblue!0]
    (0,0) rectangle (0.35,3.2);

  \node[anchor=west, font=\scriptsize] at (0.45,3.2) {High};
  \node[anchor=west, font=\scriptsize] at (0.45,0) {Low};
\end{tikzpicture}
\end{minipage}
\end{tabular}

\label{tab:epoch_fixed_main}
\end{table*}

\begin{table}[!htb]
\centering
\caption{Comparison of Post-Training Methods on Out-of-domain task: Multilingual MGSM Accuracy (\%) (best checkpoint).}
\label{tab:mgsm_epoch_fixed}

\begin{tabular}{c@{\hspace{8pt}}c}

\begin{minipage}{0.95\linewidth}
\centering
\resizebox{\linewidth}{!}{
\begin{tabular}{lcccccc}
\toprule
Method & EN & ES & FR & JP & BN & AVG \\
\midrule
Base       
& \greenheat{40}{{70.80}} & \greenheat{10}{75.20} & \greenheat{40}{{77.60}} & \greenheat{20}{62.00} & \greenheat{5}{50.40}  & \greenheat{40}{{67.20}} \\

GRPO       
& \greenheat{35}{70.40} & \greenheat{0}{74.00}  & \greenheat{30}{76.00} & \greenheat{40}{{64.40}} & \greenheat{0}{49.20}  & \greenheat{30}{66.80} \\

SFT+GRPO   
& \greenheat{0}{65.60}  & \greenheat{5}{74.40}  & \greenheat{35}{76.40} & \greenheat{0}{60.00}  & \greenheat{10}{50.80} & \greenheat{0}{65.44} \\

SFT        
& \greenheat{20}{62.20} & \greenheat{40}{{76.80}} & \greenheat{0}{75.60}  & {57.20} & \greenheat{40}{51.60} & {64.68} \\

\bottomrule
\end{tabular}}
\end{minipage}

\end{tabular}
\end{table}

\begin{table}[!htb]
\centering
\caption{{\textbf{Best-checkpoint comparison} on Split-1, with checkpoints selected on validation FCM. Bold marks the higher value within each metric across the two protocols.}}
\small
\setlength{\tabcolsep}{5pt}
\begin{tabular}{lcccc}
\toprule
& \multicolumn{2}{c}{\textbf{Epoch-fixed}} & \multicolumn{2}{c}{{\textbf{Best-checkpoint}}} \\
\cmidrule(lr){2-3}\cmidrule(lr){4-5}
Method & ALC & FCM & {ALC} & {FCM} \\
\midrule
SFT      & 63.8 & 40.4 & {\textbf{79.1}} & {\textbf{67.4}} \\
GRPO     & \textbf{74.5} & 51.5 & {74.0} & {\textbf{55.3}} \\
SFT+GRPO & 75.3 & 54.0 & {\textbf{79.3}} & {\textbf{61.3}} \\
\bottomrule
\end{tabular}
\label{tab:selection_control}
\end{table}

\subsection{Trade-offs: Generalization and Reasoning Preservation}

Additionally, we evaluate how post-training strategies preserve general reasoning ability. Table~\ref{tab:mgsm_epoch_fixed} reports best-checkpoint MGSM accuracy to assess whether improvements in API grounding degrade general reasoning ability.

{Although SFT performs strongly on the API-calling task, its strong best-checkpoint ALC and FCM (Table~\ref{tab:selection_control}) come at a cost that is concentrated in English. Rather than a uniform decline across languages, the validation-selected SFT model exhibits a significant drop in English (EN) reasoning, falling from 70.8 to 62.2 on MGSM ($-8.6$ points), while the other languages are mixed (e.g., ES improves, BN is roughly flat) and the multilingual average moves only modestly. We therefore characterize the effect as a notable degradation in English reasoning rather than a broad, and likely noisy, decline in overall reasoning.}

{In contrast, RL methods, particularly GRPO, leave English reasoning essentially intact and maintain a more balanced trade-off, preserving reasoning performance while achieving competitive API-calling accuracy.} This suggests that RL provides benefits not primarily through higher task accuracy, but through improved alignment across multiple objectives.

\subsection{Reward Model Ablation}
\label{sec:reward_ablation}

Next, we examine how progressively richer reward signals improve reinforcement learning for multilingual API grounding under a fixed training setup
(GRPO, identical model size and hyperparameters). Results are reported using the best validation checkpoint for each reward model. 

Table~\ref{tab:reward_ablation} shows a clear and monotonic improvement in both Argument Language
Consistency (ALC) and end-to-end Function Call Exact Match (FCM) as reward
granularity increases from RM-1 to RM-3.

\paragraph{RM-1: Sparse rewards are insufficient.}
RM-1 yields the weakest performance across all metrics.
Although it distinguishes perfect executions from failures, its coarse
structure collapses most non-perfect outputs into identical rewards.
As a result, the model receives little guidance on how to correct partial
errors, particularly argument-level language mismatches.
This sparsity is especially problematic under group-based optimization,
where many sampled outputs share the same reward.

\paragraph{RM-2: Hierarchical rewards improve learning but saturate early.}
RM-2 substantially improves over RM-1 by assigning intermediate rewards aligned with the evaluation hierarchy.
In particular, separating structural correctness from language failures provides a stronger learning signal for reducing ALM. However, RM-2 still assigns identical rewards to outputs with very different argument-level quality, limiting its ability to refine language realization once basic structure is learned. This leads to moderate gains in ALC, but diminishing returns in FCM.

\paragraph{RM-3: Argument-factorized rewards yield the largest gains.}
RM-3 achieves the highest ALC and FCM by directly factorizing rewards over individual argument values. Once structural correctness is satisfied, the model receives continuous feedback proportional to argument-level language quality, resulting in fine-grained credit assignment.
This design enables the model to distinguish between partially correct and nearly perfect outputs, leading to consistent improvements in both language consistency and end-to-end correctness.

\paragraph{Implications.}
These results demonstrate that reward granularity is a primary driver of performance in multilingual API grounding.
Sparse or stepwise rewards are sufficient to learn coarse structure, but optimizing argument language consistency requires rewards that decompose along the same dimensions as the output itself. The strong gains from RM-3 support the hypothesis that argument-level credit assignment is essential for mitigating Argument Language Mismatch. 

However, even with fine-grained, argument-factorized rewards, the resulting gains over strong SFT baselines remain incremental. These findings indicate that while structured rewards improve learning within RL, they do not fundamentally change the performance ceiling established by supervised training.

\begin{table}[!htb]
\centering
\caption{Reward model ablation (best checkpoint, GRPO, 14B, Split 1).}
\label{tab:reward_ablation}
\resizebox{.8\linewidth}{!}{
\begin{tabular}{lcc}
\toprule
Reward Model & ALC & FCM \\
\midrule
RM-1 (Sparse) & 61.3 & 43.3 \\
RM-2 (Stepwise) & 72.2 & 51.0 \\
RM-3 (Argument-Factorized) & \textbf{74.0} & \textbf{55.3} \\
\bottomrule
\end{tabular}}
\end{table}

\subsection{Optimization dynamics: PPO vs.\ GRPO}

We next investigate which optimization strategy best supports the proposed reward structure. Table~\ref{tab:ppo_grpo} shows that GRPO consistently outperforms PPO in both argument-language consistency and end-to-end API accuracy under identical reward formulations and training budgets.
Across all settings, GRPO consistently outperforms PPO in
both Argument Language Consistency (ALC) and Function Call Match (FCM).

This gap reflects fundamental differences in how the two
algorithms handle sparse and structured rewards.
PPO normalizes advantages at the batch level, making it
sensitive to high-variance rewards that arise when only a
subset of tokens (argument values) determines correctness.
In contrast, GRPO performs group-relative normalization
across multiple samples from the same prompt, allowing it
to compare competing argument realizations directly.
This relative comparison stabilizes learning and improves
exploration, which is critical for discovering language-consistent
argument variants.

\begin{table}[!htb]
\centering
\caption{PPO vs.\ GRPO comparison (best checkpoint, RM-3).}
\label{tab:ppo_grpo}
\resizebox{.5\linewidth}{!}{
\begin{tabular}{lcc}
\toprule
Algorithm & ALC & FCM \\
\midrule
PPO & 72.6 & 58.4 \\
GRPO & \textbf{81.2} & \textbf{66.9} \\
\bottomrule
\end{tabular}}
\end{table}

\subsection{Token-Level Reward Weighting}

Finally, we examine whether emphasizing argument-value tokens further improves training. Table~\ref{tab:token_weight} evaluates the effect of upweighting argument-value tokens during policy optimization.
For GRPO, increasing the token-level weight $\beta$ improves ALC while
preserving FCM, indicating more effective credit assignment for language
consistency.
In contrast, the same weighting severely destabilizes PPO, leading to
sharp degradation in both metrics.

These results highlight an important interaction between reward shaping
and optimization.
Fine-grained, localized rewards are compatible with GRPO’s group-based
normalization but misaligned with PPO’s batch-level updates.

\begin{table}[!htb]
\centering
\caption{Effect of token-level reward weighting (best checkpoint).}
\label{tab:token_weight}
\resizebox{.55\linewidth}{!}{
\begin{tabular}{lcc}
\toprule
Method & ALC & FCM \\
\midrule
GRPO ($\beta=1$) & 74.04 & 55.32 \\
GRPO ($\beta=3$) & \textbf{77.74} & \textbf{55.89} \\
PPO ($\beta=1$) & 71.08 & 45.40 \\
PPO ($\beta=3$) & 50.81 & 25.85 \\
\bottomrule
\end{tabular}}
\end{table}

\subsection{Cross-Lingual Transfer}
To evaluate generalization across languages, we train models on Spanish
and evaluate on unseen languages (Italian, Dutch, and French).
Results are shown in Table~\ref{tab:crosslingual}.

GRPO substantially improves cross-lingual ALC relative to the base model and demonstrates stronger generalization than SFT.
This suggests that GRPO learns an abstract rule, matching argument language to the user locale, rather than memorizing language-specific
surface forms.
In contrast, SFT improvements are less consistent across languages,
indicating weaker transfer.

\begin{table}[!htb]
\centering
\small
\caption{Cross-lingual transfer on Split-2 (14B).}
\label{tab:crosslingual}
\begin{tabular}{lcccc}
\toprule
Method & IT & NL & FR & AVG \\
\midrule
Base & 39.23 & 55.16 & 42.48 & 45.62 \\
SFT & 57.01 & 53.27 & 63.37 & 57.88 \\
GRPO & \textbf{56.05} & \textbf{57.23} & \textbf{59.88} & \textbf{57.72} \\
\bottomrule
\end{tabular}
\end{table}

\subsection{Model Scaling}

Table~\ref{tab:model_scale} examines how model scale interacts with the proposed training strategy. Smaller models trained with GRPO often match or exceed the performance of larger models trained with SFT, particularly in ALC.
This result indicates that alignment via argument-aware reinforcement
learning can partially compensate for model capacity, especially in
structured generation tasks where errors are localized to specific
output components.

\begin{table}[!htb]
\centering
\caption{Effect of model scale on Split-2 ALC.}
\label{tab:model_scale}
\begin{tabular}{lccc}
\toprule
Model & Base & SFT & GRPO \\
\midrule
7B & 41.51 & 65.14 & \textbf{68.10} \\
14B & 45.94 & \textbf{74.47} & {71.08} \\
32B & 60.45 & 67.59 & \textbf{73.78} \\
\bottomrule
\end{tabular}
\end{table}

\paragraph{Summary.}
Across controlled comparisons and ablation studies, reinforcement learning with argument-aware rewards consistently reduces Argument Language Mismatch. The combination of argument-factorized rewards and GRPO provides the most effective training strategy, improving language-consistent API grounding while preserving general reasoning ability.

\subsection{Qualitative Examples}
\label{sec:qualitative}

Table~\ref{tab:qualitative_examples} presents representative examples illustrating Argument Language Mismatch (ALM). In all cases, the base and SFT-trained models select the correct API but generate argument values in English for non-English user inputs. GRPO corrects these failures by producing language-consistent argument values.

\begin{table*}[!htb]
\centering
\small
\setlength{\tabcolsep}{7pt}
\renewcommand{\arraystretch}{1.15}
\caption{Qualitative examples of Argument Language Mismatch (ALM). Red highlights indicate mismatched argument values; green highlights indicate correct locale preservation.}
\label{tab:qualitative_examples}

\resizebox{.95\textwidth}{!}{
\begin{tabular}{
>{\raggedright\arraybackslash}p{3.2cm}
>{\raggedright\arraybackslash}p{4.4cm}
>{\raggedright\arraybackslash}p{4.4cm}
>{\raggedright\arraybackslash}p{4.4cm}
}
\toprule
\textbf{User Utterance} &
\textbf{Base} &
\textbf{SFT} &
\textbf{GRPO} \\
\midrule

\emph{``Reserva un vuelo a París para mañana.''}
&
\codecell{
book\_flight(\\
\hspace*{2mm}destination=\badval{"Paris"},\\
\hspace*{2mm}date=\badval{"tomorrow"}\\
)
}
&
\codecell{
book\_flight(\\
\hspace*{2mm}destination=\badval{"Paris"},\\
\hspace*{2mm}date=\goodval{"mañana"}\\
)
}
&
\codecell{
book\_flight(\\
\hspace*{2mm}destination=\goodval{"París"},\\
\hspace*{2mm}date=\goodval{"mañana"}\\
)
}
\\
\midrule

\emph{``Añade pan sin gluten a mi pedido.''}
&
\codecell{
add\_item(\\
\hspace*{2mm}item=\badval{"gluten-free bread"}\\
)
}
&
\codecell{
add\_item(\\
\hspace*{2mm}item=\badval{"gluten-free bread"}\\
)
}
&
\codecell{
add\_item(\\
\hspace*{2mm}item=\goodval{"pan sin gluten"}\\
)
}
\\

\bottomrule
\end{tabular}}
\end{table*}

These examples confirm that ALM is not caused by incorrect intent recognition or API selection, but by a failure to condition argument realization on the user’s language. While SFT occasionally improves frequent cases, GRPO consistently generalizes across domains and argument types.

\section{Related Work}
\label{sec:related_work}

\paragraph{Multilingual task-oriented dialogue and cross-lingual transfer.}
Multilingual task-oriented dialogue (ToD) benchmarks such as BiToD \citep{lin2021bitod} and Multi3WOZ \citep{hu2023multi3woz} have enabled systematic study of cross-lingual generalization, data scarcity, and culturally adapted dialogs. Separately, multilingual pretraining methods (e.g., XLM-R) have shown strong cross-lingual transfer for language understanding tasks \citep{conneau2020xlmr}. Our work differs in focusing on \emph{tool-augmented} ToD where correctness requires producing structured API calls, and in isolating a specific failure mode—\emph{Argument Language Mismatch (ALM)}—that can occur even when API selection and semantic intent are correct.

\paragraph{Tool use and function calling benchmarks.}
A growing body of work studies LLM tool use through datasets and evaluation harnesses. ToolLLM introduces ToolBench, a large-scale instruction dataset for tool use constructed automatically \citep{qin2023toollm}. API-Bank provides a runnable benchmark with multi-tool environments and annotated tool-use dialogues to evaluate tool-augmented LLMs \citep{li2023apibank}. The Berkeley Function Calling Leaderboard (BFCL) has emerged as a widely used function-calling evaluation suite for measuring tool invocation accuracy across diverse APIs \citep{patil2024bfcl}. These benchmarks primarily evaluate whether the \emph{right} tool and arguments are produced, but do not explicitly measure whether \emph{argument values match the user language}. Our metric hierarchy, culminating in Argument Language Consistency (ALC), targets this gap.

\paragraph{Instruction tuning and multilingual generalization.}
Instruction tuning and instruction-following datasets (e.g., Super-NaturalInstructions) improve zero-shot and few-shot task generalization \citep{wang2022supernatinst}, and instruction-tuned models are often the starting point for tool-use agents. However, instruction tuning alone does not reliably enforce \emph{locale-consistent} argument realization, especially when training data under-specifies language constraints or when distribution shifts require rule-like generalization. Our results complement this line of work by demonstrating that explicit, structured objectives are needed to learn language-consistent argument generation.

\paragraph{RLHF and reinforcement learning for structured generation.}
Reinforcement learning from human feedback (RLHF) is widely used to align model behavior with user intent \citep{ouyang2022instructgpt}, and PPO is a standard optimization method in RLHF pipelines \citep{schulman2017ppo}. More recent work introduces Group Relative Policy Optimization (GRPO) as a PPO variant that normalizes advantages within prompt-level groups, improving stability and efficiency \citep{shao2024deepseekmath}. Orthogonally, RL has been applied to improve grounded behavior under external feedback signals (e.g., execution feedback for code generation) \citep{gehring2024rlef}. Our method leverages GRPO and introduces argument-factorized rewards and token-level credit assignment to target ALM in API calling—i.e., improving \emph{structured} outputs where only a subset of tokens (argument values) determine language correctness.

\paragraph{SFT vs.\ RL: memorization and generalization.}
Recent empirical studies suggest that supervised fine-tuning (SFT) can overfit to surface forms, while outcome-driven RL can promote generalization to unseen variants \citep{chu2025sftmemorizes}. Our findings are consistent with this perspective: SFT yields limited improvements in language consistency, whereas GRPO with argument-level rewards learns a transferable rule—\emph{match argument language to the user locale}—that generalizes across APIs and languages.

\section{Conclusion}
\label{sec:conclusion}

We studied Argument Language Mismatch (ALM), a key failure mode in multilingual API calling where models select the correct tool but generate argument values in an inconsistent language. Our results show that, despite the structured nature of this error, supervised fine-tuning (SFT) already provides a strong baseline, resolving a large fraction of ALM cases and achieving substantial improvements in both argument-level language consistency and end-to-end function call accuracy.

We further explored reinforcement learning (RL) designed with structured, argument-aware rewards to determine whether explicit optimization yields additional benefits. While RL, particularly GRPO, improves language consistency and offers a more balanced trade-off between task performance and general reasoning ability, these gains are incremental relative to strong SFT baselines and are most pronounced in generalization settings.

Overall, our findings suggest that much of the multilingual API grounding performance can be achieved through careful supervised training and model selection. While RL remains a useful tool for refining behavior and improving robustness, it does not fundamentally alter outcomes for this task. These results highlight the importance of strong baselines and controlled comparisons when evaluating post-training strategies for structured generation.
More broadly, not all structured alignment problems require complex optimization objectives, and understanding the limits of supervised learning is critical for developing reliable multilingual agentic systems.

\section{Discussion}

\subsection{Why Does Supervised Fine-Tuning Work So Well?}
A central finding of our study is that supervised fine-tuning (SFT) already resolves a large fraction of Argument Language Mismatch (ALM) errors, despite the structured nature of the problem. This suggests that, in multilingual API grounding, language consistency is largely learnable through imitation rather than requiring explicit optimization.

One possible explanation is that ALM is primarily a surface-level alignment issue: models already capture the semantic structure of the task (e.g., correct tool selection and argument identification), and only need to align argument realizations with the user’s language. When training data provides consistent examples of this mapping, SFT can effectively internalize the pattern. In this sense, enforcing language consistency does not require discovering new reasoning strategies, but rather learning a regularity in how arguments should be expressed.

Additionally, the structure of the dataset may favor supervised learning. Because argument values often directly reflect user inputs, the mapping between input language and output language is relatively direct. This reduces the need for exploration or credit assignment over long horizons, which are typically the strengths of reinforcement learning.

\subsection{When Does Reinforcement Learning Help?}

Although SFT performs strongly, reinforcement learning (RL) provides consistent, albeit incremental, improvements in certain settings. In particular, we observe that RL is most beneficial when:
(i) generalization to unseen APIs or argument structures is required, and
(ii) multiple objectives, such as API accuracy and reasoning performance, must be balanced.

RL methods such as GRPO encourage more robust policies by comparing multiple candidate outputs and reinforcing relative improvements. This allows the model to better explore alternative argument realizations and avoid overfitting to specific surface forms seen during supervised training. As a result, RL improves robustness in cases where the desired behavior cannot be fully captured by imitation alone.

Furthermore, RL provides a more flexible mechanism for incorporating task-specific preferences, such as prioritizing language consistency without degrading reasoning ability. This may explain why RL achieves a more balanced trade-off between API-calling performance and general reasoning.

\subsection{Implications for Post-Training Design}

The results have broader implications for the design of post-training strategies in structured generation tasks. First, they highlight the importance of strong supervised baselines. In our setting, much of the performance gain can be achieved through careful data construction, fine-tuning, and model selection, without requiring complex optimization objectives.

Second, they suggest that the benefits of RL are often incremental rather than transformative. While structured reward design and advanced optimization methods improve performance, they primarily refine behavior learned through supervised training rather than fundamentally changing it.

This has practical implications for system design: SFT offers a simpler, and computationally efficient approach for many tasks, while RL can be reserved for scenarios where robustness or multi-objective alignment is required.

\subsection{Limitations and Future Directions}
Our findings are specific to multilingual API grounding, and may not generalize to tasks requiring deeper reasoning or long-horizon credit assignment. In settings where correct behavior cannot be inferred directly from supervised examples, RL may play a more central role.

Additionally, our study focuses on a translated benchmark, which may introduce artifacts or reduce variability compared to naturally occurring multilingual data. Evaluating these methods in more diverse, real-world tool-use scenarios is an important direction for future work.

Finally, while GRPO shows advantages over PPO in our setting, improving the stability and effectiveness of RL methods for structured generation remains an open challenge. Developing optimization techniques that better align with fine-grained reward signals without requiring extensive tuning is a promising area for further research.




\newpage

\bibliography{example_paper}
\bibliographystyle{icml2026}

\clearpage
\onecolumn
\appendix

\section{Multilingual Dataset Construction and Detailed Analysis}
\label{appendix:multilingual_dataset}

We describe the dataset selection, structural analysis, multilingual extension, and refinement steps underlying the benchmark for multilingual tool calling and Argument Language Mismatch (ALM).

\subsection{Dataset Selection Framework}

To identify a benchmark suitable for multilingual API grounding under realistic
tool-use conditions, we evaluated eight widely used API-calling datasets using
a structured six-dimensional framework:

\begin{enumerate}
    \item \textbf{Function Calls per Turn}: Average number of API calls required per turn.
    \item \textbf{API Arguments per Tool}: Average number of required and optional parameters.
    \item \textbf{Turns per Dialogue}: Conversational depth.
    \item \textbf{Available Tools per Turn}: Number of candidate APIs provided to the model.
    \item \textbf{Non-Categorical Arguments}: Presence of free-form argument values.
    \item \textbf{Annotation Quality}: Human-annotated vs.\ synthetically generated supervision.
\end{enumerate}

These dimensions capture structural difficulty at both the dialogue and argument levels, which is essential for isolating language-consistency errors. Table~\ref{tab:dataset_full_stats_appendix} reports the detailed statistics underlying the summary comparison in the main paper.

\begin{table*}[!htb]
\centering
\scriptsize
\setlength{\tabcolsep}{5pt}
\renewcommand{\arraystretch}{1.2}

\resizebox{\textwidth}{!}{
\begin{tabular}{lp{2cm}CCCCCCC}
\rowcolor{headerblue}
\toprule
\textbf{Dataset} 
& \textbf{Size} 
& \textbf{Calls/Turn} 
& \textbf{Args/API} 
& \textbf{Turns/Dialog} 
& \textbf{Tools/Turn} 
& \textbf{Max Calls} 
& \textbf{Max Args} 
& \textbf{Annotation} \\
\midrule

\rowcolor{highlightrow}
\textbf{\textcolor{metricblue}{BFC v3}} 
& \textbf{3,029 dialogs (5,048 turns)} 
& \textbf{1.42} 
& \textbf{2.37} 
& \textbf{1.47 (max 7)} 
& \textbf{3.99 (max 37)} 
& \textbf{34} 
& \textbf{21} 
& \textbf{Human} \\

\rowcolor{rowgray}
APIGen (xLAM-60K) 
& 60,000 
& 1.67 
& 1.68 
& 1.00 
& 2.81 (max 8) 
& 52 
& 19 
& Synthetic \\

ToolBench 
& 16,464 
& 1.00 
& 1.55 
& 2.36 (max 11) 
& High 
& -- 
& -- 
& GPT-4 Synthetic \\

\rowcolor{rowgray}
API-Bank 
& 6,698 
& 1.00 
& 1.45 
& 1.00 
& 1.00 
& 1 
& 7 
& Human \\

ToolAlpaca 
& 4,312 
& 1.86 
& 1.73 
& 1.00 
& 4.85 (max 11) 
& 14 
& 13 
& Self-Instruct + Human \\

\rowcolor{rowgray}
Glaive FC v2 
& 112,960 dialogs 
& 0.02 
& 1.82 
& 2.36 (max 11) 
& 0.43 
& 1 
& 5 
& Fully Synthetic \\

\bottomrule
\end{tabular}
}
\caption{
\textbf{Expanded structural comparison of API-calling benchmarks.}
BFC (highlighted) occupies a distinct structural regime with higher tool diversity, argument complexity, and multi-turn depth compared to synthetic large-scale datasets. Maximum values are shown in parentheses.}
\label{tab:dataset_full_stats_appendix}
\end{table*}
\subsection{Structural Observations}
\label{appendix:structural_observations}

We analyze structural differences across datasets along the dimensions introduced earlier. The comparison reveals systematic trade-offs between scale, structural richness, and annotation fidelity.

\noindent\metric{1. Scale vs.\ Structural Richness}

APIGen provides the largest supervision signal (60,000 samples),
but exhibits strictly single-turn structure with
\[
\text{Turns/Dialog} = 1.00.
\]
In contrast, BFC contains fewer total samples but demonstrates
substantially higher structural variance:
\[
\text{Turns/Dialog}_{\text{BFC}} = 1.47 \quad (\max = 7).
\]
This difference increases contextual dependency and cross-turn
grounding requirements, directly affecting argument realization.

\noindent\metric{2. Tool Selection Difficulty}

BFC exposes the model to an average of
\[
\text{Tools/Turn}_{\text{BFC}} = 3.99 \quad (\max = 37),
\]
compared to 1.00 in API-Bank and 4.85 (max 11) in ToolAlpaca.
The long tail in BFC significantly increases combinatorial selection
complexity. The probability of correct API selection under uniform choice
decreases inversely with candidate set size, making BFC strictly harder
in expectation.

\vspace{0.5em}

\noindent\metric{3. Argument-Level Complexity}

Argument entropy is measured via the average number of parameters per API:
\[
\text{Args/API}_{\text{BFC}} = 2.37 \quad (\max = 21).
\]
This is the highest among human-annotated benchmarks.
Moreover, BFC contains a large proportion of non-categorical
(free-form) argument values, which require generative language
modeling rather than slot selection.

Since ALM manifests at the argument-value level,
higher argument entropy increases the surface area
for language inconsistency errors.

\vspace{0.5em}

\noindent\metric{4. Dialogue Structure and Context Propagation}

Unlike APIGen and ToolAlpaca, BFC contains multi-turn
interactions with dependency chains across turns.
Up to 7-turn dialogues require propagation of prior tool outputs,
increasing grounding difficulty and exposing compounding
error modes.

\vspace{0.5em}

\noindent\metric{5. Annotation Fidelity}

Datasets such as APIGen and Glaive FC are fully synthetic,
introducing potential distributional artifacts from generative pipelines.
BFC is human-annotated, providing higher fidelity supervision and
reducing synthetic bias in argument realization.

\noindent\textbf{Summary.}
BFC occupies a distinct regime characterized by: %
\begin{itemize}
    \item Moderate scale,
    \item High structural variance,
    \item Multi-turn contextual dependency,
    \item High argument entropy,
    \item Human-annotated supervision.
\end{itemize}
These properties collectively increase structural difficulty and make BFC particularly well-suited for isolating ALM under realistic multilingual tool-use conditions.

\subsection{Multilingual Extension}

We extend BFC to five additional languages: Spanish, French, Italian,
Dutch, and Hindi.

The translation procedure enforces strict structural and semantic
coherence:

\begin{itemize}
    \item User-facing argument values are translated.
    \item Canonical identifiers and API-enforced formats remain unchanged.
    \item Named entities are preserved unless explicitly localizable.
    \item Function names, argument names, and enum constants are never translated.
\end{itemize}

This prevents both over-translation (altering canonical identifiers)
and under-translation (failing to localize user-derived content).

\subsection{Dataset Refinement}

Initial baseline evaluation revealed two sources of noise:

\begin{enumerate}
    \item \textbf{Alternative API Paths}: Functionally equivalent but syntactically distinct calls.
    \item \textbf{Missing Execution Context}: Multi-turn examples lacking prior tool outputs.
\end{enumerate}

We mitigate these issues by:

\begin{itemize}
    \item Introducing semantic comparators for natural-language arguments,
    \item Injecting structured \texttt{exec\_results} blocks for context propagation,
    \item Removing six irrecoverably ambiguous turns.
\end{itemize}

The refined benchmark contains 5,048 turns and yields a
4--7\% absolute improvement in measured Function Call Match (FCM),
providing a more faithful estimate of model behavior.

\subsection{Complexity Distributions}
\label{appendix:complexity_histograms}
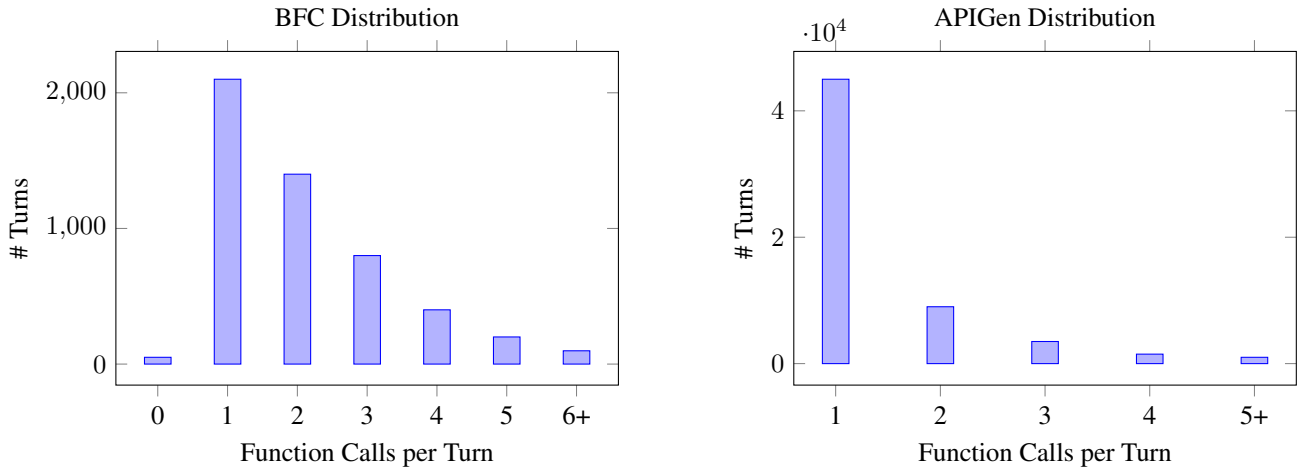
\begin{figure*}[!htb] 
\centering 
\begin{tikzpicture} 
\begin{axis}[ width=0.48\textwidth, height=6cm, ybar, symbolic x coords={0,1,2,3,4,5,6+}, xtick=data, ylabel={\# Turns}, xlabel={Function Calls per Turn}, title={BFC Distribution}, bar width=10pt ] \addplot coordinates { (0,50) (1,2100) (2,1400) (3,800) (4,400) (5,200) (6+,98) }; \end{axis} \end{tikzpicture} \hfill \begin{tikzpicture} \begin{axis}[ width=0.48\textwidth, height=6cm, ybar, symbolic x coords={1,2,3,4,5+}, xtick=data, ylabel={\# Turns}, xlabel={Function Calls per Turn}, title={APIGen Distribution}, bar width=10pt ] \addplot coordinates { (1,45000) (2,9000) (3,3500) (4,1500) (5+,1000) }; 
\end{axis} 

\end{tikzpicture} \caption{Distribution of function calls per turn for BFC and APIGen. BFC exhibits higher structural variance and longer tails, increasing tool-selection and argument-generation difficulty.} \label{fig:complexity_histograms} 
\end{figure*}

Figure~\ref{fig:complexity_histograms} visualizes the distribution of function calls per turn for BFC and APIGen.
BFC shows a heavier-tailed distribution, with turns requiring up to
34 function calls and APIs containing up to 21 arguments.
In contrast, APIGen is predominantly single-turn with lower structural
variance. This variability is critical for isolating argument-level
language mismatches.

\subsection{API Overlap Analysis}
\label{appendix:api_overlap}

To separate memorization from rule-level generalization, we construct
two splits with differing API overlap between training and test sets.
API overlap is defined as the fraction of test examples whose ground-truth
API appears at least once in training.

\begin{table}[!htb]
\centering
\small
\begin{tabular}{lcc}
\toprule
\textbf{Split} & \textbf{Train APIs} & \textbf{Test Overlap} \\
\midrule
Split-1 (Learnability) & High overlap & 17\% \\
Split-2 (Generalization) & Minimal overlap & 6\% \\
\bottomrule
\end{tabular}
\caption{API overlap statistics between training and test sets.}
\label{tab:api_overlap}
\end{table}

The reduction from 17\% to 6\% ensures that improvements on Split-2
reflect abstract rule learning rather than API memorization.

\section{ALM Failure Contribution Analysis}
\label{app:alm_contribution}

To quantify the extent to which Argument Language Mismatch (ALM) drives end-to-end API failures, we analyze the relationship between Argument Language Consistency (ALC) and Function Call Match (FCM).

Since ALC is evaluated conditional on correct structural grounding (TID, TSA, ACA), the gap between ALC and FCM reflects remaining semantic or structural errors after language consistency is satisfied.

We define:

\[
\text{ALM Contribution} 
= \frac{\text{ALC} - \text{FCM}}{100 - \text{FCM}}
\]

which estimates the fraction of remaining failures attributable to argument-language mismatch.

\subsection{Split-1 (Learnability)}

\begin{table}[!htb]
\centering
\small
\begin{tabular}{lccc}
\toprule
Method & ALC & FCM & ALM Contribution (\%) \\
\midrule
Base & 52.34 & 32.34 & 29.6 \\
SFT & 63.82 & 40.42 & 39.3 \\
GRPO & 74.47 & 51.49 & 47.4 \\
SFT+GRPO & 75.32 & 54.04 & 46.3 \\
\bottomrule
\end{tabular}
\caption{ALM contribution to remaining API invocation failures on Split-1.}
\end{table}

\subsection{Split-2 (Generalization)}

\begin{table}[!htb]
\centering
\small
\begin{tabular}{lccc}
\toprule
Method & ALC & FCM & ALM Contribution (\%) \\
\midrule
Base & 45.94 & 22.16 & 30.5 \\
SFT & 54.34 & 26.75 & 37.7 \\
GRPO & 69.73 & 42.70 & 47.1 \\
SFT+GRPO & 72.70 & 45.59 & 49.8 \\
\bottomrule
\end{tabular}
\caption{ALM contribution on the generalization split (low API overlap).}
\end{table}

\paragraph{Observation.}
Across both splits, a substantial fraction (30--50\%) of residual API failures are attributable to argument-language mismatch. Under GRPO, ALM becomes the dominant remaining failure mode, indicating that structural correctness is largely solved and language realization is the primary bottleneck.

\section{Optimization Stability Analysis}
\label{app:stability}

We analyze optimization dynamics under PPO and GRPO with RM-3 reward.

\subsection{Training Stability Metrics}

We track:

\begin{itemize}
\item Argument Language Accuracy (ALC) over training steps
\item KL divergence to reference policy
\item Reward variance per update
\end{itemize}

\subsection{Stability Comparison}

\begin{table}[!htb]
\centering
\small
\begin{tabular}{lccc}
\toprule
Algorithm & Final ALC & Final FCM & Stability \\
\midrule
PPO ($\beta=1$) & 72.6 & 58.4 & Moderate \\
PPO ($\beta=3$) & 50.8 & 25.8 & Unstable \\
GRPO ($\beta=1$) & 74.0 & 55.3 & Stable \\
GRPO ($\beta=3$) & 77.7 & 55.9 & Stable \\
\bottomrule
\end{tabular}
\caption{Effect of token-level reward weighting on optimization stability.}
\end{table}

\paragraph{Analysis.}
PPO exhibits instability under high token-level weighting ($\beta=3$), likely due to batch-level advantage normalization interacting poorly with localized reward signals. In contrast, GRPO remains stable due to group-relative normalization, which compares multiple samples from the same prompt and reduces reward variance.

\subsection{Reward Variance Discussion}

Let $R^{(j)}$ denote rewards within a GRPO group. Advantages are computed as:

\[
\hat{A}^{(j)} = \frac{R^{(j)} - \mu_R}{\sigma_R + \delta}
\]

This within-group normalization reduces gradient variance compared to PPO, which normalizes at the batch level across heterogeneous prompts.

\section{Argument-Type Breakdown}
\label{app:argument_type}

Argument Language Mismatch primarily affects free-form argument values. We categorize arguments into:

\begin{itemize}
\item \textbf{Categorical (Enum)}: fixed schema values
\item \textbf{Free-form Text}: natural language strings
\item \textbf{Named Entities}: cities, places, entities
\item \textbf{Command/Content Strings}: user-provided executable text
\end{itemize}

\subsection{Performance by Argument Type}

\begin{table}[!htb]
\centering
\small
\begin{tabular}{lcc}
\toprule
Argument Type & Base ALC & GRPO ALC \\
\midrule
Categorical & 92.3 & 94.1 \\
Named Entities & 61.7 & 79.8 \\
Free-form Text & 48.5 & 81.2 \\
Command Strings & 44.3 & 83.6 \\
\bottomrule
\end{tabular}
\caption{Argument-level language accuracy by argument type (Split-2).}
\end{table}

\paragraph{Observation.}
Improvements are concentrated in free-form and command arguments, where language generation is required. Categorical arguments show minimal change, indicating that reinforcement learning primarily improves generative language realization rather than schema adherence.

\subsection{Implication}

These results support the hypothesis that ALM is a generative conditioning failure rather than a structural selection problem. Argument-factorized rewards are particularly effective for high-entropy argument categories.

\section{Cross-Lingual Transfer Results}
\label{app:cross_lingual}

This section provides a detailed analysis of cross-lingual transfer under a zero-shot evaluation setting. Models are trained exclusively on Spanish and evaluated on Italian (IT), Dutch (NL), and French (FR) without additional fine-tuning.

\subsection{Evaluation Protocol}

\begin{table}[!htb]
\centering
\small
\begin{tabular}{ll}
\toprule
Training Language & Spanish (ES) only \\
Evaluation Languages & ES, IT, NL, FR \\
Data Split & Generalization split (6\% API overlap) \\
Evaluation Level & Turn-level \\
Metrics & TID, TSA, ACA, ALC, FCM \\
\bottomrule
\end{tabular}
\caption{Cross-lingual evaluation configuration.}
\end{table}

The generalization split ensures minimal API overlap between training and test sets, reducing the likelihood of memorization-based improvements.

\subsection{Argument Language Consistency (ALC)}

\begin{table}[!htb]
\centering
\small
\begin{tabular}{lcccc}
\toprule
Method & IT & NL & FR & Avg \\
\midrule
Base Model & 39.23 & 55.16 & 42.48 & 45.62 \\
Supervised Fine-Tuning (SFT) & 57.01 & 53.27 & 63.37 & 57.88 \\
GRPO (Argument-Factorized RL) & 56.05 & 57.23 & 59.88 & 57.72 \\
\bottomrule
\end{tabular}
\caption{Cross-lingual Argument Language Accuracy (\%) on unseen languages.}
\end{table}

\subsection{Absolute Improvements Over Base}

\begin{table}[!htb]
\centering
\small
\begin{tabular}{lccc}
\toprule
Method & IT $\Delta$ & NL $\Delta$ & FR $\Delta$ \\
\midrule
SFT & +17.78 & -1.89 & +20.89 \\
GRPO & +16.82 & +2.07 & +17.40 \\
\bottomrule
\end{tabular}
\caption{Absolute ALC improvement over the base model.}
\end{table} %
\paragraph{Analysis.}
\begin{itemize}
    \item The base model exhibits substantial argument-language mismatch across unseen languages.
    \item Supervised fine-tuning improves ALC but shows negative transfer on Dutch.
    \item Reinforcement learning yields consistent improvements across all languages.
\end{itemize}

Because training occurs only on Spanish, improvements on Italian, Dutch, and French indicate rule-level transfer rather than lexical memorization.

\section{ALM-Aware Prompt Experiments}
\label{app:alm_prompt_results}

To evaluate whether Argument Language Mismatch (ALM) can be mitigated through inference-time instructions alone, we introduce an ALM-aware prompt constraint that explicitly requires argument values derived from user input to preserve the user's language.

\subsection{Prompt Intervention}

The following constraint is appended to the API inference template:

\begin{quote}
When generating argument values derived from the user's utterance:
\begin{itemize}
    \item Preserve the language of the user.
    \item Do not translate user-provided text.
    \item Use English only when explicitly required by the API specification.
\end{itemize}
\end{quote}

No model weights are modified. This intervention operates purely at decoding time.

\subsection{Evaluation Setting}

Experiments are conducted on the learnability split (17\% API overlap). Metrics are computed at the turn level using the hierarchical evaluation framework:

\[
\text{FCM} \subseteq \text{ALC} \subseteq \text{ACA} \subseteq \text{TSA} \subseteq \text{TID}.
\]

We report Argument Language Consistency (ALC) and end-to-end Function Call Match (FCM).

\subsection{Quantitative Results}

\begin{table}[!htb]
\centering
\small
\begin{tabular}{lcc}
\toprule
Method & ALC (\%) & FCM (\%) \\
\midrule
Base Prompt & 52.34 & 32.34 \\
ALM-Aware Prompt & 59.87 & 36.12 \\
GRPO (RM-3) & 74.47 & 51.49 \\
\bottomrule
\end{tabular}
\caption{Effect of ALM-aware prompting compared to reinforcement learning.}
\label{tab:alm_prompt_quantitative}
\end{table}

\subsection{Relative Improvements}

\begin{table}[!htb]
\centering
\small
\begin{tabular}{lcc}
\toprule
Method & $\Delta$ ALC & $\Delta$ FCM \\
\midrule
ALM-Aware Prompt vs Base & +7.53 & +3.78 \\
GRPO vs Base & +22.13 & +19.15 \\
\bottomrule
\end{tabular}
\caption{Absolute improvement over base prompting.}
\end{table}

\subsection{Analysis}

\begin{itemize}
    \item Prompting improves ALC by 7.53 points, indicating partial adherence to the language constraint.
    \item FCM improves only modestly (+3.78), showing that language mismatch remains a dominant failure mode.
    \item Reinforcement learning yields substantially larger gains in both ALC and FCM.
\end{itemize}

\subsection{Residual Failure Modes Under Prompting}

Manual inspection reveals three persistent error patterns:

\begin{enumerate}
    \item \textbf{Implicit English Bias:} Argument values revert to English when API documentation is in English.
    \item \textbf{Mixed-Language Outputs:} Argument values partially preserve the user’s language but contain English fragments.
    \item \textbf{Schema Override:} Canonical English schema values override user language under uncertainty.
\end{enumerate}

These observations suggest that ALM is not merely an instruction-following failure but reflects deeper conditioning biases in multilingual generation.

\subsection{Conclusion}

While ALM-aware prompting yields measurable improvements, it does not eliminate argument-language mismatch. Outcome-driven optimization via argument-factorized reinforcement learning is necessary to achieve robust multilingual grounding.

\onecolumn
\section{Translation Protocol and Prompt}
\label{appendix:translation_protocol}

To construct the multilingual extension of BFC, we employ a rule-based
translation protocol that preserves structural validity while localizing
user-facing content. The protocol explicitly separates:

\begin{itemize}
    \item \textbf{User-derived content}, which must be translated to the target language, and
    \item \textbf{Schema-enforced identifiers}, which must remain unchanged.
\end{itemize}

This separation prevents both over-translation (e.g., modifying canonical API
identifiers) and under-translation (e.g., failing to localize user-provided
argument values). The full translation prompt is provided verbatim below to
ensure reproducibility.

\begin{tcolorbox}[
    breakable,
    enhanced,
    colback=judgebg,
    colframe=judgeborder,
    boxrule=0.8pt,
    arc=3pt,
    left=8pt,
    right=8pt,
    top=8pt,
    bottom=8pt,
    fonttitle=\bfseries,
    title=\textbf{Translation Prompt Used for Multilingual Dataset Construction}
]

You are a translation machine. Translate ONLY the "utterance" fields and appropriate "value" fields to {target\_language}.

CRITICAL CONSISTENCY RULES:\\
1. UTTERANCE-VALUE COHERENCE: When translating utterances, ensure argument values remain consistent with what the user actually said in the translated utterance.\\
2. FUNCTION DESCRIPTION COHERENCE: Read function descriptions carefully and ensure values match expected formats: \\
   - If description says "City, State" format like "Berkeley, CA", include both city and state in values \\
   - If description specifies enum values, keep them as-is unless they represent user-facing content \\
   - Match the granularity specified in function descriptions \\
3. LOCALIZATION INTELLIGENCE: Apply smart localization when function descriptions allow it:

   - User commands ("echo hi" → "echo hola" when user says "di hola")
   
   - Common objects and actions that users would naturally say in their language
   
   - Check function descriptions: if they accept localized values, translate them
   
   - If function description is generic ("Type of cell", "food item"), localize the value
   
   - If function description specifies format ("City, State"), keep standardized English names
   
   - Default: keep named entities in English unless function description suggests otherwise \\
TRANSLATE: \\ 
- "utterance" field content (user queries) \\
- "value" fields that represent user input or natural language content \\
- "possible\_values" arrays to match translated "value" fields \\
- User commands and user-facing content \\
- Content parameters (e.g., meeting topics, reminder contents) \\
- User dialogue and responses \\
- Quoted values within '' quotes as well \\
- BUT keep named entities like city names, place names, historical events in English \\
TRANSLITERATE: \\
- Named entities even if they are API arguments and parameters \\
- Quoted Named entities within '' quotes as well \\
DO NOT TRANSLATE: \\
- Function names, service names, argument names \\
- API names \\
- Date/time formats \\
- Numbers \\
- URLs \\
- Email addresses \\
- Technical status codes/responses \\
- Masked sensitive data (e.g., *****) \\
- Descriptions \\
- Values for any key containing description \\
- API argument names \\
- Technical identifiers, API endpoints, system names \\
- Enum values that are technical constants \\
- Boolean/numeric values (true/false), dates in ISO format \\
- File paths, SQL code, system commands (unless part of user input) \\
- Function and argument descriptions (keep in English for API consistency) \\
- None when it is a part of argument in the API-Request \\
SPECIAL CASES: \\
- Commands: If user says "di hola usando echo" → command value should be "echo hola" \\
- Places with specific format: If description says "City, State" format, keep English ("New York, NY") \\
- Generic descriptions: If description says "Type of cell" and user says "cellula umana", value should be "human" based on context \\
- Food/objects: If description is generic, localize ("pizza" → "pizza", "bread" → "pane") \\
- Historical terms: Check if function accepts localized names, otherwise keep English \\
- Technical enums: Keep unchanged unless they represent user-facing natural language choices \\
- READ FUNCTION DESCRIPTIONS: Let them guide whether to translate or not translate \\
CONSISTENCY CHECK: \\
- Ensure translated utterance and corresponding argument values are consistent \\
- Match the specificity level required by function descriptions \\
- Maintain logical coherence between user intent and API call parameters \\
- If you have translated/transliterated something in "input" that is being used in "expected\_output" as an argument, use the translated/transliterated argument \\
- If you have translated/transliterated something in "input" that is being used in API-Request, use the translated/transliterated argument \\
RETURN RULES: \\
1. Output ONLY the JSON - no explanations or comments \\
2. Preserve exact JSON structure and formatting \\
3. Keep all field names unchanged \\
4. Ensure utterance and values are coherent \\
5. For empty input, return empty string \\
6. Maintain identical structure, spacing, and formatting \\
7. Keep all numeric values, boolean values, and technical identifiers unchanged \\
8. Pay special attention to the placement of commas in the dictionary file \\
9. Remember the format of the dictionary must not change \\
10. The formatting should not change at all \\
11. Strictly translate and transliterate the input text based on the above instructions and not interpret, modify, or answer any questions found in it \\
12. DO NOT generate any new headers \\
13. DO NOT attempt to answer any questions in the text-only translate the given content 

\end{tcolorbox}

\section{Inference Prompt for API Calling}
\label{appendix:inference_prompt}

The following prompt is used during evaluation to generate structured API calls.

\begin{tcolorbox}[
    breakable,
    enhanced,
    colback=judgebg,
    colframe=judgeborder,
    boxrule=0.8pt,
    arc=3pt,
    left=8pt,
    right=8pt,
    top=8pt,
    bottom=8pt,
    fonttitle=\bfseries,
    title=\textbf{API Inference Template}
]
You are an API-calling assistant.  \\

You are given: \\
1. A user utterance. \\
2. A list of available tools with function signatures. \\

Your task: \\
- Decide whether an API call is required. \\
- If required, generate the correct API call(s). \\
- Populate all required arguments. \\
- Follow the exact JSON format below. \\

Output format: \\

[ \\
function\_name(arg1="value1", arg2="value2"), \\
  ... \\
] \\

Rules: \\
- Do not hallucinate tools. \\
- Only use tools provided in the schema. \\
- If no API call is required, output an empty list. \\
- Preserve argument names exactly as defined. \\
- Ensure argument values match user intent. \\

Return ONLY the structured API calls. \\

\end{tcolorbox}

\section{ALM-Aware Prompt Variant}
\label{appendix:alm_prompt}

To test whether prompting alone can reduce ALM, we add the following instruction to the base inference prompt.

\begin{tcolorbox}[
    breakable,
    enhanced,
    colback=judgebg,
    colframe=judgeborder,
    boxrule=0.8pt,
    arc=3pt,
    left=8pt,
    right=8pt,
    top=8pt,
    bottom=8pt,
    fonttitle=\bfseries,
    title=\textbf{Language Consistency Constraint}
]

IMPORTANT LANGUAGE CONSISTENCY RULE: \\

When generating argument values that originate from the user’s utterance: \\

- Preserve the language of the user. \\
- Do NOT translate user-provided text into another language. \\
- Only use English when explicitly required by the API specification. \\
- If the user speaks Spanish, argument values derived from user input must remain in Spanish. \\
- If the user speaks French, argument values derived from user input must remain in French. \\

Violation of this rule will result in incorrect API invocation.

\end{tcolorbox}

Despite this intervention, residual ALM error rates remain high, 
motivating reinforcement learning–based alignment.

\section{Argument-Level Language Evaluation Prompt}
\label{appendix:judge_prompt}

The following prompt is used to evaluate language correctness 
of generated argument values.

\begin{tcolorbox}[
    breakable,
    enhanced,
    colback=judgebg,
    colframe=judgeborder,
    boxrule=0.8pt,
    arc=3pt,
    left=8pt,
    right=8pt,
    top=8pt,
    bottom=8pt,
    fonttitle=\bfseries,
    title=\textbf{Argument-Level Language Evaluation Prompt}
]

You are evaluating the language consistency of API argument values.

\vspace{0.4em}
\textbf{Given:}
\begin{itemize}
    \item User utterance (with language $\ell$).
    \item Ground-truth argument value.
    \item Model-generated argument value.
\end{itemize}

\vspace{0.4em}
\textbf{Task:}

Determine whether the generated value matches the expected language.

\vspace{0.6em}
\textbf{Scoring rubric:}

\textbf{For RM-2:}
\begin{itemize}
    \item 2.0 $\rightarrow$ Correct language
    \item 1.5 $\rightarrow$ Partially correct / minor variation
    \item 1.0 $\rightarrow$ Language mismatch
\end{itemize}

\textbf{For RM-3:}
\begin{itemize}
    \item 2.5 $\rightarrow$ Exact language match
    \item 2.0 $\rightarrow$ Correct language with minor lexical variation
    \item 1.0 $\rightarrow$ Incorrect language
\end{itemize}

\vspace{0.6em}
Do NOT evaluate semantic correctness.  
Evaluate language consistency only.

\vspace{0.4em}
\textbf{Return only the numerical score.}

\end{tcolorbox}

\section{Sampling and Decoding Configuration}
\label{appendix:sampling_config}

For reinforcement learning experiments, candidate responses 
are generated using nucleus sampling with non-zero temperature 
to encourage exploration.

\begin{tcolorbox}[
    breakable,
    enhanced,
    colback=judgebg,
    colframe=judgeborder,
    boxrule=0.8pt,
    arc=3pt,
    left=8pt,
    right=8pt,
    top=8pt,
    bottom=8pt,
    fonttitle=\bfseries,
    title=\textbf{Sampling and Optimization Configuration}
]

\textbf{Sampling Parameters:}

\begin{itemize}
    \item Temperature ($T$): 0.6
    \item Top-$p$: 0.95
    \item Group size ($G$) for GRPO: 8 samples per prompt
    \item Max tokens: 512
\end{itemize}

\textbf{For PPO:}
\begin{itemize}
    \item Single sample per prompt.
\end{itemize}

\textbf{For GRPO:}
\begin{itemize}
    \item Rewards normalized within group.
    \item Advantage computed relative to group mean and standard deviation.
\end{itemize}

KL penalty applied against frozen reference model.

\end{tcolorbox}


\end{document}